\documentclass[letterpaper]{article} 
\usepackage{aaai2026}  
\usepackage{times}  
\usepackage{helvet}  
\usepackage{courier}  
\usepackage[hyphens]{url}  
\usepackage{graphicx} 
\usepackage{natbib}  
\usepackage{caption} 
\usepackage{booktabs}

\pubnote{} 
\copyrighttext{} 

\usepackage{algorithm}
\usepackage{algorithmic}
\usepackage{amssymb}
\usepackage{amsmath}
\usepackage{xcolor} 
\usepackage{subcaption}

\usepackage{newfloat}
\usepackage{listings}
\DeclareCaptionStyle{ruled}{labelfont=normalfont,labelsep=colon,strut=off} 
\floatstyle{ruled}
\newfloat{listing}{tb}{lst}{}
\floatname{listing}{Listing}
\title{Ctx2TrajGen: Traffic Context-Aware Microscale Vehicle Trajectories using Generative Adversarial Imitation Learning}

\author {
    Joobin Jin\textsuperscript{\rm 1},
    Seokjun Hong\textsuperscript{\rm 1},
    Gyeongseon Baek\textsuperscript{\rm 1},
    Yeeun Kim\textsuperscript{\rm 2},
    Byeongjoon Noh\textsuperscript{\rm 3}\thanks{Corresponding author}
}
\affiliations {
    \textsuperscript{\rm 1}Department of Future Convergence Technology, Soonchunhyang University, Asan, 31538, South Korea\\
    \textsuperscript{\rm 2}
    Department of Civil and Environmental \& Construction Engineering, University of Central Florida, Orlando 32816, USA\\
    \{yeeun.kim\}@ucf.edu
    \textsuperscript{\rm 3}Department of AI and Big Data, Soonchunhyang University, Asan, 31538, South Korea\\
    \{jjb0821, hongsj, baekgs011109,  powernoh\}@sch.ac.kr
}

\usepackage{bibentry}

\begin{document}

\maketitle

\begin{abstract}
Precise modeling of microscopic vehicle trajectories is critical for traffic behavior analysis and autonomous driving systems. We propose \textbf{Ctx2TrajGen}, a context-aware trajectory generation framework that synthesizes realistic urban driving behaviors using GAIL. Leveraging PPO and WGAN-GP, our model addresses nonlinear interdependencies and training instability inherent in microscopic settings. By explicitly conditioning on surrounding vehicles and road geometry, Ctx2TrajGen generates interaction-aware trajectories aligned with real-world context. Experiments on the drone-captured DRIFT dataset demonstrate superior performance over existing methods in terms of realism, behavioral diversity, and contextual fidelity, offering a robust solution to data scarcity and domain shift without simulation. \textbf{The source code will be made publicly available upon paper acceptance}.
\end{abstract}


\section{Introduction}Precise modeling of microscopic vehicle trajectories plays a crucial role in traffic behavior analysis and the development of autonomous driving systems \cite{sun2017discovering, qi2024microscopic}. High-resolution trajectory data that capture frame-level spatial and dynamic states—such as position, velocity, and acceleration—enable detailed analyses of fine-grained driving behaviors, including abrupt lane changes, evasive maneuvers, and close-range interactions \cite{naing2024fine, lu2025dual}. Such data serve as foundational input for traffic safety diagnostics, driver behavior modeling, and high-fidelity simulation \cite{yan2024evaluation}, and offer distinct research value by incorporating contextual and interactional information that cannot be obtained from GPS-based macroscale data \cite{yan2023learning}.

However, acquiring these high-resolution datasets remains technically and institutionally challenging. The deployment of roadside sensors, aerial drone operations, and compliance with privacy regulations all impose significant limitations on the scalability of spatiotemporal data collection \cite{sun2024cudc, xiong2025multi}. Furthermore, unlike macroscale trajectories that are largely goal-directed, microscopic trajectories are characterized by \textit{nonlinear interdependencies} that emerge from their sensitivity to surrounding vehicles and road context \cite{lu2022vehicle, wang2025c2f}, making them particularly difficult to model accurately.

To address these challenges, we propose \textbf{Ctx2TrajGen}, a novel vehicle trajectory generation method designed to synthesize realistic microscopic driving behaviors under real-world conditions. Built upon Generative Adversarial Imitation Learning (GAIL) \cite{ho2016generative}, the proposed Ctx2TrajGen learns directly from expert demonstrations without requiring handcrafted reward functions. To overcome training instability and enhance policy expressiveness—challenges arising from nonlinear interdependencies—we integrate Proximal Policy Optimization (PPO) \cite{schulman2017proximal} and Wasserstein GAN with Gradient Penalty (WGAN-GP) \cite{gulrajani2017improved} into the learning process.

A key feature of Ctx2TrajGen is its explicit conditioning of the policy on the temporally evolving states of surrounding vehicles and lane-level road structure. By embedding both interaction cues and environmental context—including dynamic agents and road geometry—into the decision process, the model generates context-sensitive and interaction-aware driving sequences that reflect the behavioral diversity observed in urban traffic.

Extensive experiments on the real-world drone-captured DRIFT dataset \cite{lee2025drift} show that Ctx2TrajGen outperforms state-of-the-art baselines in terms of realism, diversity, and contextual alignment. The framework successfully mitigates domain shift issues without relying on simulation, and ablation studies confirm the contribution of each architectural component. This work addresses the dual challenge of data scarcity and unstable policy learning, establishing a new foundation for microscopic behavior generation grounded in real traffic contexts.
\section{Related Work}Recent work on vehicle trajectory generation has largely focused on macroscale mobility modeling. These approaches typically generate origin-destination (OD) sequences based on GPS traces for applications in urban planning and traffic flow simulation \cite{cabanas2025human, owais2024deep, rong2023goddag}. For instance, TrajVAE \cite{chen2021trajvae} utilized an LSTM encoder-decoder with a variational autoencoder (VAE) to generate diverse OD-level trajectories, while TrajGDM \cite{chu2024simulating} applied diffusion probabilistic models to simulate human mobility with enhanced behavioral diversity. To improve spatial fidelity, \citet{jiang2023continuous} introduced a Two-Stage GAN that progressively refines OD paths into road-aligned trajectories, and \citet{zhu2023difftraj} proposed DiffTraj to improve statistical consistency over GAN-based methods using traffic-level metrics. TrajGAIL \cite{choi2021trajgail} applied GAIL to generate realistic OD-level vehicle paths by imitating expert trajectories in urban networks.

While these methods demonstrate advances in generative modeling, they remain limited to coarse-grained trajectories and overlook the rich contextual dependencies found in microscopic driving behavior. Lane-level interactions, road geometry, and temporal coordination among vehicles are rarely addressed. Additionally, many of these methods rely on simulated environments for training and validation, limiting their applicability in complex real-world scenarios.

In contrast, our work focuses on the generation of microscopic trajectories with a high degree of behavioral realism and contextual sensitivity. By integrating adversarial imitation learning with explicit spatiotemporal conditioning and recurrent policy structures, our framework advances trajectory generation toward interaction-aware modeling in real-world traffic environments.

\section{Preliminary}\subsection{Problem Definition}
The primary objective of Ctx2TrajGen is to generate microscale position sequences that resemble real vehicle travel trajectories observed in urban road environments. The notion of “similarity” between the generated and real trajectories can be defined from two perspectives. First, dataset-level similarity assesses the statistical or distributional resemblance across the entire trajectory dataset. Second, trajectory-level similarity evaluates the physical plausibility and situational realism of vehicle states and maneuvers within each generated trajectory. These two perspectives collectively define the goal of producing realistic and diverse microscale trajectories.

One of the critical aspects to consider in microscale vehicle trajectory generation is that driving behavior is strongly influenced by interactions with surrounding vehicles. For instance, when the leading vehicle moves slowly or when adjacent lanes are congested, the following vehicle has no choice but to reduce its speed and follow the leading vehicle unless it executes a lane change maneuver. Such spatiotemporal interactions directly shape the trajectory of each vehicle and introduce a complex decision-making structure that cannot be easily explained by the plans of a single agent. While it is theoretically possible to formulate these interactions using a partially observable Markov decision process (POMDP) \cite{choi2021trajgail}, in practice, doing so often leads to an intractably large state space and a highly complex policy structure. Therefore, microscale trajectory generation requires a learning-based approach that effectively incorporates dynamic and localized traffic context, including relative positions, speeds, the distribution of surrounding vehicles, and lane geometry.

To address these challenges, this study adopts a modeling approach that begins with the design of context-aware state representations, capable of capturing localized road and vehicle interaction information. Specifically, we construct a state space that reflects spatiotemporal interactions with surrounding vehicles and define a Markov decision process (MDP) based on this representation. Subsequently, we employ a GAIL framework to learn interaction-aware driving policies from real-world urban traffic data collected via drone-based aerial imagery, enabling the generation of diverse and realistic microscale trajectories grounded in actual traffic behavior.

\subsection{Formulation}
A vehicle trajectory is defined as a temporally ordered sequence of two-dimensional positions over discrete time frames, representing the path of an individual vehicle in a planar space. In this study, the trajectory of a vehicle over a time horizon of \(h\) frames is denoted as \(\textit{Traj} = \{(x_1, y_1), (x_2, y_2), \dots, (x_h, y_h)\}\). Among all vehicles, we designate one as the \textit{ego vehicle}, for which the goal is to simulate a plausible driving trajectory under realistic urban traffic conditions. 

To achieve this, we define the agent’s state at each time step \(t\), \(\mathbf{s}_t\), as a context-aware representation that encapsulates not only the ego vehicle’s internal dynamics but also its external driving context. Specifically, the ego state vector \(\mathbf{z}_t = [x_t, y_t, v_{x,t}, v_{y,t}, a_{x,t}, a_{y,t}]\) describes the vehicle’s kinematic attributes along both spatial axes. To capture interaction cues, we incorporate a feature matrix \(\mathbf{V}_t \in \mathbb{R}^{N \times d}\) that encodes the relative motion of \textit{surrounding vehicles}. Here, \(N\) denotes the number of surrounding vehicles, while \(d\) indicates the dimensionality of features, encompassing positions and velocities relative to the ego. In addition, a one-hot lane occupancy vector \(\boldsymbol{\ell}_t\) indicates the ego vehicle’s current lane position, providing road-level structural information. This composite state representation \(\mathbf{s}_t = (\mathbf{z}_t, \mathbf{V}_t, \boldsymbol{\ell}_t)\) is designed to enable the policy to make context-sensitive decisions that reflect both spatial interactions and roadway constraints.

Based on this formulation, we model the trajectory generation task as an MDP, formalized as four variables: \((\mathcal{S}, \mathcal{A}, T, R)\). Here, \(\mathcal{S}\) denotes the state space that the agent operates in, and \(\mathcal{A}\) is the action space, where each action \(\mathbf{a}_t = [\Delta x_t, \Delta y_t]\) represents the ego vehicle’s displacement in the two-dimensional plane. The transition function \(T(\cdot)\) deterministically updates the next state (\(\mathbf{s}_{t+1} \in \mathcal{S}\)) given the current state \(\mathbf{s}_t \in \mathcal{S}\) and action \(\mathbf{a}_t \in \mathcal{A}\), and the reward function \(R(\mathbf{s}_t, \mathbf{a}_t)\) evaluates the desirability of taking action \(\mathbf{a}_t\) in state \(\mathbf{s}_t\). A policy \(\pi_{\theta}\) is a function that maps each state to an action. This MDP serves as the foundation for learning the policy.

In Ctx2TrajGen, the policy \(\pi_{\theta}\) is learned through adversarial imitation, where the agent generates trajectories by interacting with the MDP environment, aiming to mimic expert driving behaviors without relying on handcrafted reward functions.

\section{Proposed Method}\begin{figure*}[!t]
    \centering
    \includegraphics[width=0.9\linewidth]{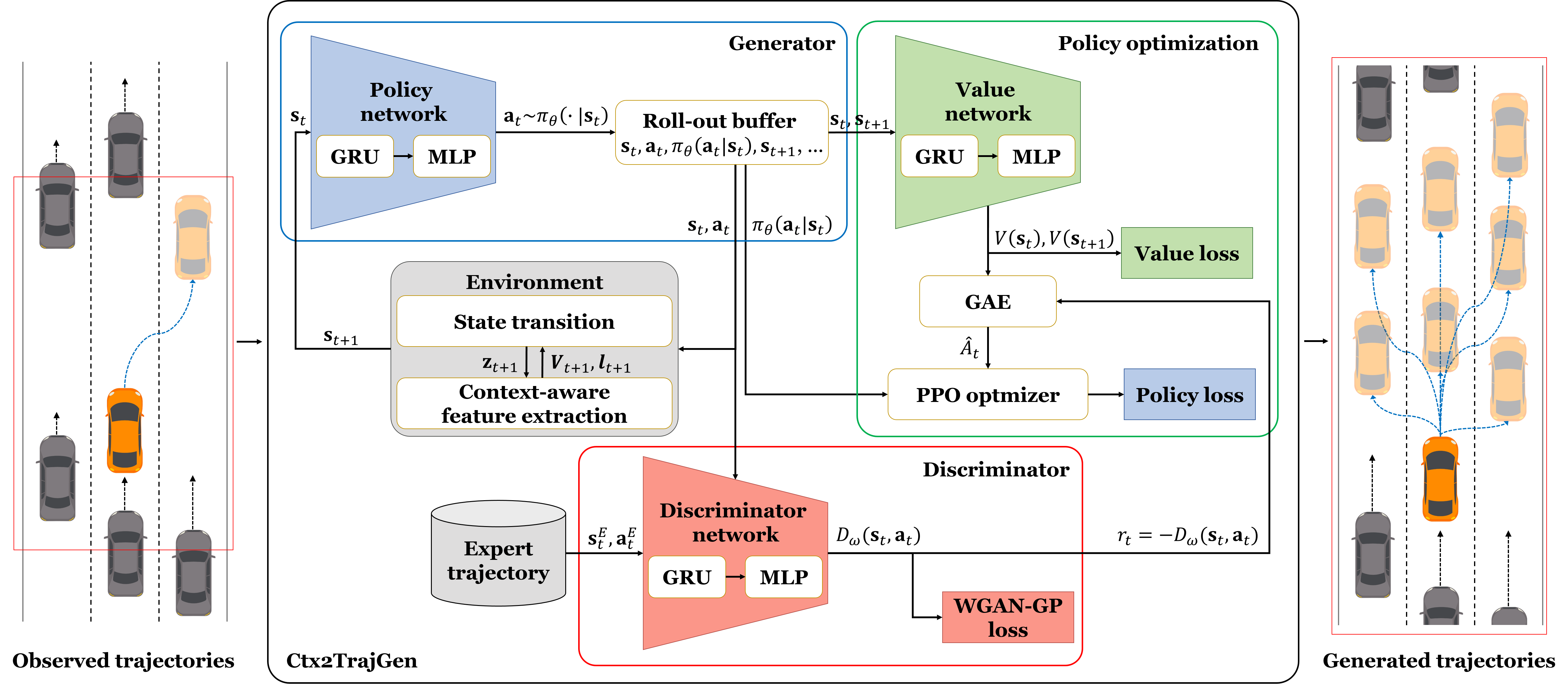}
    \caption{Overall architecture of \textbf{Ctx2TrajGen}. The proposed framework is built upon the GAIL paradigm and is designed to generate microscale vehicle trajectories that reflect realistic interactions with surrounding agents and contextual traffic information. While maintaining the core structure of a generator and a discriminator, our model incorporates a WGAN-GP loss in the discriminator to enhance training stability, and adopts PPO for policy learning by jointly optimizing the advantage-based policy loss and value loss. Additionally, each network is equipped with a GRU backbone to effectively capture temporal dependencies. This architecture distinguishes itself from vanilla GAIL by improving both training robustness and representational capacity.}
    \label{fig:fig-architecture}
\end{figure*}

The overall architecture of the proposed \textbf{Ctx2TrajGen} framework is depicted in Figure~\ref{fig:fig-architecture}, built upon the GAIL and comprising three core components: an environment that simulates state transitions based on vehicle dynamics and local traffic context, a generator implemented as a policy network that outputs actions from observed states, and a discriminator that differentiates expert trajectories from those generated by the policy using a loss function. The entire system is trained through an adversarial process, where the generator is optimized using PPO algorithm to enhance training stability and the discriminator is stabilized with WGAN-GP objective, supporting convergence toward realistic and diverse microscale trajectory generation.

\subsection{Context-Aware Encoded Environment}
The environment in the proposed Ctx2TrajGen is designed to provide a deterministic and structured simulation space that enables the policy to interact with context-rich observations and generate realistic vehicle trajectories. It plays two primary roles: updating the ego vehicle’s kinematic state in response to actions, and extracting a context-aware representation that encapsulates the traffic dynamics surrounding the agent.

At each time step \(t\), the agent samples an action \( \mathbf{a}_t = [\Delta x_t, \Delta y_t] \) from its policy, which is applied to the ego vehicle’s current state to produce the next state via a deterministic state transition function. The state transition is defined as a function \( T: (\mathbf{s}_t, \mathbf{a}_t) \mapsto \mathbf{s}_{t+1} \), which updates the ego vehicle's position and velocity based on its motion model, without any stochastic noise or external perturbations. This supports that generated motions remain physically consistent and spatially coherent across time.

\begin{figure}[!t]
    \centering
    \includegraphics[width=1.0\linewidth]{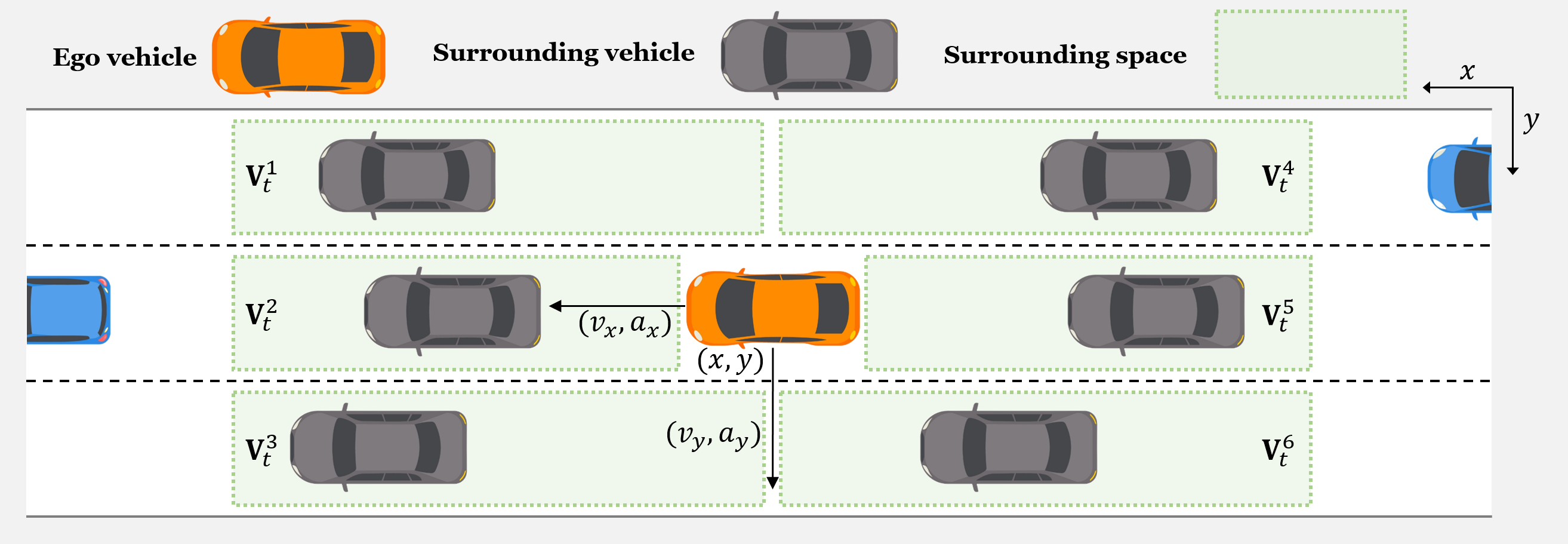}
    \caption{Ego and surrounding vehicle feature encoding. This figure depicts the ego vehicle's position, velocity, and acceleration, surrounding vehicles' relative positions and velocities encoded in \(\mathbf{V}_t\), and lane occupancy via \(\boldsymbol{\ell}_t\), centered for context-aware representation in Ctx2TrajGen.}
    \label{fig:fig-ego-vehicle}
\vspace{-10pt}
\end{figure}

After the transition, the environment assembles a full observation \( \mathbf{s}_{t+1} \) by integrating the updated ego state with interaction-aware features derived from nearby traffic agents. Specifically, the environment employs a fixed spatial template centered on the ego vehicle to identify a set of surrounding vehicles and encodes their relative positions and velocities into a matrix representation. This process captures localized traffic context while maintaining spatial symmetry and temporal continuity. Figure~\ref{fig:fig-ego-vehicle} illustrates the spatial configuration used for selecting surrounding vehicles in our implementation.

The resulting state \(\mathbf{s}_{t+1} \) encapsulates both self-centric motion cues and mesoscopic traffic context, which collectively inform the policy’s decision-making process. This environment structure allows the agent to adaptively react to heterogeneous traffic scenarios, such as car-following, lane changes, and gap acceptance, without relying on explicit rule-based modeling.

\subsection{Policy Generator}
The trajectory generator in Ctx2TrajGen is implemented as a stochastic policy network that sequentially produces driving actions conditioned on temporally grounded and context-aware observations. It constitutes the core module that translates observed states into multimodal driving behaviors, while preserving temporal coherence throughout the trajectory. The generator interacts with the environment to simulate future states and accumulates experience in a structured memory module for downstream optimization.

\subsubsection{Policy Network}
To learn time-dependent and context-sensitive maneuvers, the policy network adopts a Gated Recurrent Unit (GRU) as its core encoder, which recursively processes the input state sequence and maintains a hidden state that captures historical driving context. The hidden representation is subsequently passed through fully connected layers that parameterize a Gaussian Mixture Model (GMM) distribution over actions. This architecture facilitates the policy in modeling varied and stochastic maneuvers such as lane changes, car-following, and overtaking, with \(K\) mixture components.

At each timestep \( t \), the network samples a two-dimensional action \( \mathbf{a}_t = [\Delta x_t, \Delta y_t] \) from the learned GMM conditioned on the hidden state \( \mathbf{h}_t \) derived from the encoded observation \( \mathbf{s}_t \). The resulting stochastic action model is formulated as:
\begin{equation}
\mathbf{a}_t \sim \sum_{k=1}^{K} w_t^{k} \, \mathcal{N}(\boldsymbol{\mu}_t^{k}, \boldsymbol{\sigma}_t^{k})
\end{equation}
where each component of the mixture model is defined by a weight \( w_t^{k} \), a mean vector \( \boldsymbol{\mu}_t^{k} \), and a standard deviation \( \boldsymbol{\sigma}_t^{k} \). This formulation allows the policy to generate diverse and realistic driving patterns in complex traffic environments.

\subsubsection{Roll-out Buffer}
To support stable policy learning, we introduce a roll-out buffer that systematically stores the data collected during the generator's interaction with the environment. At each time step, the buffer records the observed state \( \mathbf{s}_t \), the sampled action \( \mathbf{a}_t \), the next state \( \mathbf{s}_{t+1} \) resulting from the environment’s deterministic transition, and the log-probabilities of the action under both the current policy \( \pi_\theta \) and the previous policy \( \pi_{\text{old}} \). These records are stored as tuples and accumulated over multiple steps to form complete trajectories.

This roll-out buffer acts as an interface between the trajectory generator and subsequent modules such as the discriminator, value network, and PPO optimizer. It provides the basis for computing surrogate objectives, estimating advantage values using Generalized Advantage Estimation (GAE) \cite{schulman2015high}, and comparing policy performance across iterations. Additionally, the buffer serves as the starting point for trajectory roll-outs used during inference, ensuring that each generated sequence is grounded in previously observed interaction histories. By separating the data sampling and training phases, the buffer enhances training efficiency while preserving temporal consistency in generated trajectories.

\subsection{Trajectory Discriminator}
The discriminator in Ctx2TrajGen is designed to differentiate between expert and generated trajectories by evaluating the authenticity of state-action pairs. Unlike conventional GAIL frameworks that rely on binary cross-entropy loss, our model adopts WGAN-GP to promote stable and meaningful reward estimation.

Formally, the discriminator is parameterized as a temporal neural network \( D_\omega(\mathbf{s}_t, \mathbf{a}_t) \), which assigns a scalar score to each state-action pair. To capture temporal dependencies in sequential driving behavior, the architecture comprises a GRU encoder followed by a multilayer perceptron (MLP). The GRU processes the trajectory sequence, including concatenated state \(\mathbf{s}_t\) and action \(\mathbf{a}_t\), to encode context-aware temporal features, and the MLP maps the GRU hidden states to scalar outputs.

The training objective follows the WGAN-GP formulation, where the discriminator is trained to assign higher scores to expert trajectories and lower scores to generated ones. Let \( \mathbb{E}_{\text{exp}} \) and \( \mathbb{E}_{\text{gen}} \) denote expectations over expert and generator trajectories, respectively. The discriminator loss is defined as:
\begin{equation}
\begin{aligned}
\mathcal{L}_D 
&= \mathbb{E}_{(\mathbf{s}, \mathbf{a}) \sim \mathbb{E}_{\text{gen}}} [D_\omega(\mathbf{s}, \mathbf{a})] - \mathbb{E}_{(\mathbf{s}, \mathbf{a}) \sim \mathbb{E}_{\text{exp}}} [D_\omega(\mathbf{s}, \mathbf{a})]\\ 
&\quad + \lambda \, \mathbb{E}_{\hat{\mathbf{x}} \sim \mathbb{P}_{\hat{\mathbf{x}}}} [(\|\nabla_{\hat{\mathbf{x}}} D_\omega(\hat{\mathbf{x}})\|_2 - 1)^2]
\end{aligned}
\end{equation}
where \( \mathbb{P}_{\hat{\mathbf{x}}} \) is the distribution of interpolated samples between expert and generated trajectories, and \( \lambda \) is the gradient penalty coefficient.

The output score \( D_\omega(\mathbf{s}_t, \mathbf{a}_t) \) is transformed into the reward for the policy network, following the inverse reinforcement learning paradigm to minimize the Wasserstein distance from expert trajectories:
\begin{equation}
r_t = -D_\omega(\mathbf{s}_t, \mathbf{a}_t)
\end{equation}

This design enables the discriminator to provide rich and smooth reward signals that guide the policy toward expert-like behaviors, while maintaining training stability through gradient regularization. 

\subsection{Policy Optimization}
The training of the trajectory generator in Ctx2TrajGen is based on a reinforcement learning framework utilizing discriminator reward signals to enhance policy behavior. We break down the optimization into three key components: GAE, PPO, and multi-objective loss formulation.

\subsubsection{Generalized Advantage Estimation (GAE)}
To estimate the temporal credit assignment between actions and their long-term effects, we adopt GAE. It computes a smoothed estimate of the advantage function to reduce variance while maintaining bias controllability. Given the value network \( V_\psi(\cdot) \), the temporal difference residual is computed as:
\begin{equation}
\delta_t = r_t + \gamma V_\psi(\mathbf{s}_{t+1}) - V_\psi(\mathbf{s}_t)
\end{equation}
and the advantage at time \( t \) is estimated as:
\begin{equation}
A_t = \sum_{l=0}^{T-t} (\gamma \lambda)^l \delta_{t+l}
\end{equation}
where \( \gamma \in [0,1] \) is the discount factor and \( \lambda \in [0,1] \) is the smoothing parameter.

\subsubsection{PPO Algorithm}
To stably update the policy, we apply PPO \cite{schulman2017proximal}, which restricts the update magnitude by clipping the ratio between the new and old policies. Given the policy \( \pi_\theta \) and the old policy \( \pi_{\text{old}} \), the ratio is:
\begin{equation}
r_t(\theta) = \frac{\pi_\theta(\mathbf{a}_t | \mathbf{s}_t)}{\pi_{\text{old}}(\mathbf{a}_t | \mathbf{s}_t)}
\end{equation}
and the clipped surrogate objective is defined as:

\begin{equation}
\mathcal{L}_{\text{policy}} = - \mathbb{E}_t \left[ \min \left( r_t(\theta) A_t, \, \text{clip}(r_t(\theta), 1 - \epsilon, 1 + \epsilon) A_t \right) \right]
\end{equation}
where \( \epsilon \) is a predefined threshold that controls the allowable deviation from the previous policy.

\subsection{Loss Optimization}

The overall optimization objective incorporates three components: the surrogate policy loss from PPO, the value regression loss, and an entropy regularization term. The value network is trained using a mean squared error between the predicted value and the cumulative return:
\begin{equation}
\mathcal{L}_{\text{value}} = \mathbb{E}_t \left[ \left( V_\psi(\mathbf{s}_t) - V_t^{\text{target}} \right)^2 \right]
\end{equation}
with the return target defined as:
\begin{equation}
V_t^{\text{target}} = \sum_{l=0}^{T-t} \gamma^l r_{t+l}
\end{equation}

The entropy loss promotes stochastic exploration by encouraging policy uncertainty:
\begin{equation}
\mathcal{L}_{\text{entropy}} = \mathbb{E}_t \left[ \mathcal{H}(\pi_\theta(\cdot | \mathbf{s}_t)) \right]
\end{equation}

The final objective function to be minimized is:
\begin{equation}
\mathcal{L}_{\text{total}} = \mathcal{L}_{\text{policy}} + c_1 \mathcal{L}_{\text{value}} - c_2 \mathcal{L}_{\text{entropy}}
\end{equation}
where \( c_1 \) and \( c_2 \) are tunable hyperparameters. All gradient updates are performed with respect to both the policy network parameters \( \theta \) and value network parameters \( \psi \).

Reward signals \( r_t \) are derived from the output of the discriminator \( D_\omega(\mathbf{s}_t, \mathbf{a}_t) \), providing adversarial feedback that encourages the generator to imitate expert-like behaviors.
\section{Validation}\subsection{Experimental Setup}

\subsubsection{Data}
The evaluation of the Ctx2TrajGen framework relies on the DRIFT dataset \cite{lee2025drift}, a comprehensive urban traffic resource recorded by synchronized drones at approximately 250 meters altitude across nine interconnected intersections in Daejeon, South Korea. This dataset, comprising 81,699 high-resolution vehicle trajectories, offers comprehensive microscopic states, including speed, acceleration, and steering angles, derived from video synchronization. To enhance contextual fidelity, the data underwent preprocessing, focusing on site C, utilizing 920 trajectories, where lane structure information with curvature was integrated. Dynamic context was further enriched by extracting features from six surrounding vehicles using Region of Interest (RoI) mapping, supporting a robust representation of the ego vehicle’s driving environment.

\subsubsection{Baselines}
The performance of the Ctx2TrajGen model is evaluated through comparative experiments with four recognized trajectory generation models, selected based on their documented impact in published studies:

\begin{itemize}
    \item \textbf{TrajSynVAE} \cite{wang2023synthesizing}: A VAE framework that generates diverse GPS trajectories.
    \item \textbf{TrajGDM} \cite{chu2024simulating}: A stochastic diffusion model simulating macroscale human mobility.
    \item \textbf{LSTM-TrajGAN} \cite{rao2020lstm}: An LSTM-GAN hybrid designed for time-series trajectory generation.
    \item \textbf{DiffTraj} \cite{zhu2023difftraj}: A probabilistic diffusion model enhancing GPS trajectory fidelity.
\end{itemize}

\subsubsection{Implementation Details}
The Ctx2TrajGen framework is designed to enable robust generation of microscopic trajectories. The policy, value, and discriminator networks employ a two-layer GRU with 128 hidden units to model temporal dependencies. Learning rates are set to 5e-5 for the policy, 1e-4 for the value function, and 1e-8 for the discriminator, with the PPO epsilon tuned to 0.98 within a threshold range of [0.95, 1.01] to balance exploration and stability. The WGAN-GP gradient penalty coefficient is fixed at 1.0 to ensure training stability. These configurations, refined through grid search, support the model’s accurate emulation of expert driving patterns.

\subsubsection{Evaluation Metrics}
The evaluation of the Ctx2TrajGen framework employs distribution-based metrics to assess trajectory similarity. The key metrics are:

\begin{itemize}
    \item \textbf{Maximum Mean Discrepancy (MMD)}: A nonparametric measure of distributional difference.

    \item \textbf{Wasserstein Distance (WD)}: A metric quantifying divergence via transport cost.

    \item \textbf{Kullback-Leibler (KL) Divergence}: A measure of relative entropy between distributions.

    \item \textbf{Jensen-Shannon (JS) Divergence}: A symmetric KL variant for distribution comparison.

\end{itemize}

\subsection{Main Results}

The Ctx2TrajGen model, our proposed framework, achieves an MMD of 0.0021, a KL Divergence of 1.2543, and a JS Divergence of 0.2746, significantly outperforming baseline models. This represents a 99.9\% reduction from DiffTraj’s MMD of 2.000, highlighting the model’s proficiency in capturing complex traffic interactions and enhancing predictive accuracy in urban contexts. With a Wasserstein Distance of 0.2781, Ctx2TrajGen outperforms LSTM-TrajGAN’s 0.4460 but falls short of TrajSynVAE’s 0.2233. However, LSTM-TrajGAN’s higher KL value of 7.5378 and JS value of 1.0839, along with TrajSynVAE’s KL value of 21.3349 and JS value of 0.6090, indicate a narrow focus on specific metrics, potentially compromising overall trajectory coherence. Consequently, Ctx2TrajGen’s superior performance across multiple metrics, including MMD, KL, and JS, demonstrates enhanced stability, a balanced distribution, and a reduced risk of mode collapse. This refined modeling approach provides a foundation for improved traffic flow analysis and adaptive traffic signal optimization.

\begin{table}[!t]
\centering
\renewcommand{\arraystretch}{1.1}
\fontsize{9.3pt}{12pt}\selectfont
\setlength{\tabcolsep}{6pt}
\begin{tabular}{@{}lcccc@{}}
\toprule
\textbf{Model} & \textbf{MMD ↓} & \textbf{WD ↓} & \textbf{KL ↓} & \textbf{JS ↓} \\
\midrule
DiffTraj       & 2.0000 & 6.2296 & 3.2938  & 0.2792 \\
LSTM-TrajGAN   & 1.9956 & 0.4460 & 7.5378  & 1.0839 \\
TrajGDM       & 1.9715 & 5.1126 & 18.1868 & 0.5244 \\
TrajSynVAE   & 1.9873 & \textbf{0.2233} & 21.3349 & 0.6090 \\
\textbf{Ctx2TrajGen (ours)}   & \textbf{0.0021} & 0.2781 & \textbf{1.2543} & \textbf{0.2746} \\
\bottomrule
\end{tabular}
\caption{Comparison of the proposed Ctx2TrajGen model with established trajectory generation baselines, evaluated through distributional similarity metrics. Lower values signify enhanced alignment with expert trajectories.}
\label{tab:tab-quantitative-results}
\end{table}

\begin{figure}[!t]
    \centering
    \includegraphics[width=0.9\linewidth]{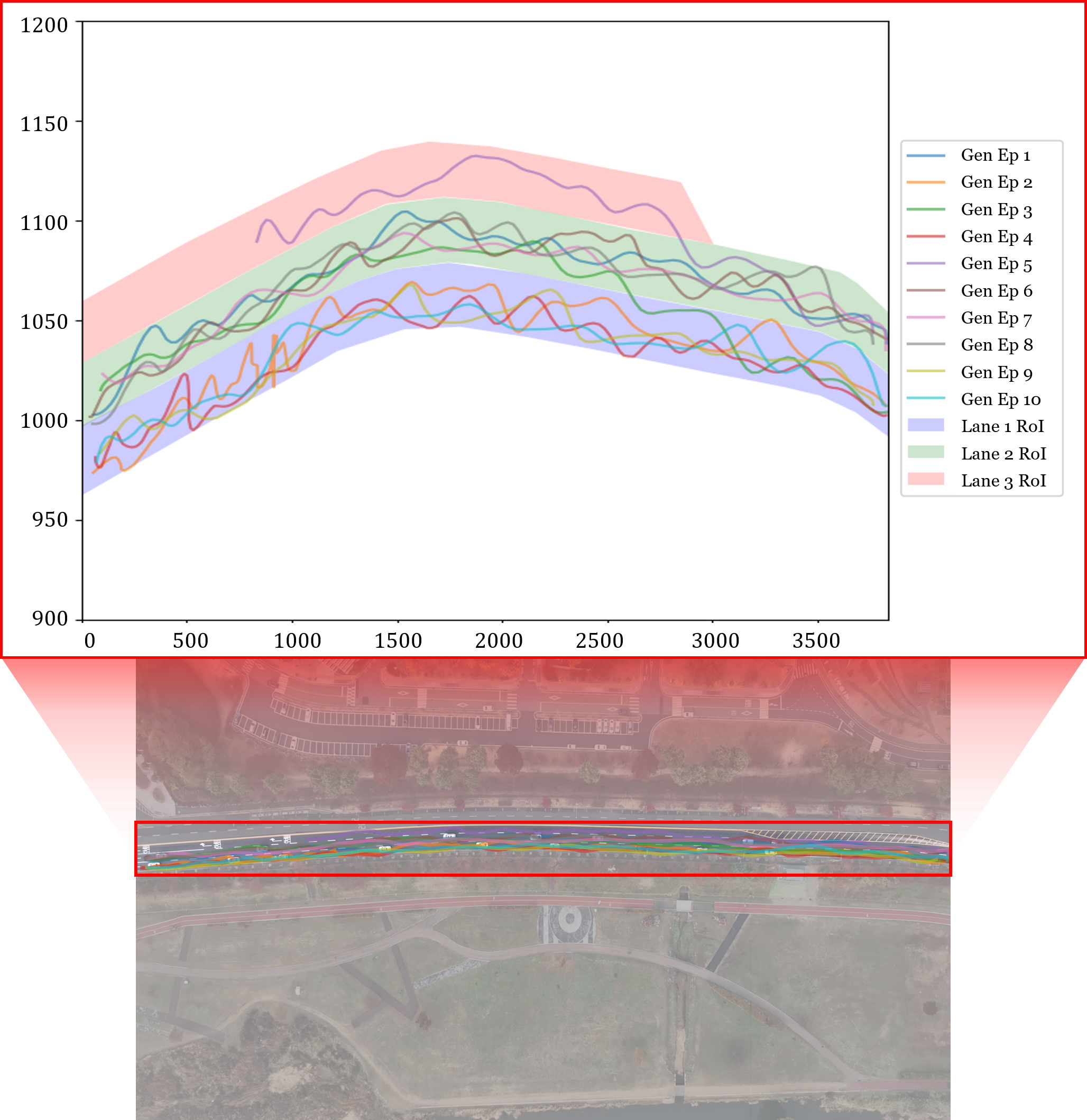}    
    \caption{Generated vehicle trajectories plotted with enlarged road details and lane-specific RoI.}
    \label{fig:fig-trajectory-generation-plot}
\end{figure}

To evaluate the generative fidelity of the Ctx2TrajGen model, we analyzed trajectory samples across diverse driving conditions, as shown in Figure~\ref{fig:fig-trajectory-generation-plot}. The model yields trajectories that ensure continuous path integrity, align with lane configurations, and mirror interactions with adjacent vehicles, demonstrating its adeptness at resolving fine spatial details. This strength supports consistent trajectory tracking across varied terrains, setting it apart from less detailed methodologies. Such precision in capturing microscopic dynamics strengthens real-time path consistency, improves path prediction accuracy, and provides a reliable foundation for dynamic traffic management.

\subsection{Ablation Studies}

\begin{figure}[!t]
    \centering
    \begin{subfigure}[b]{0.9\columnwidth}
        \centering
        \includegraphics[width=\linewidth]{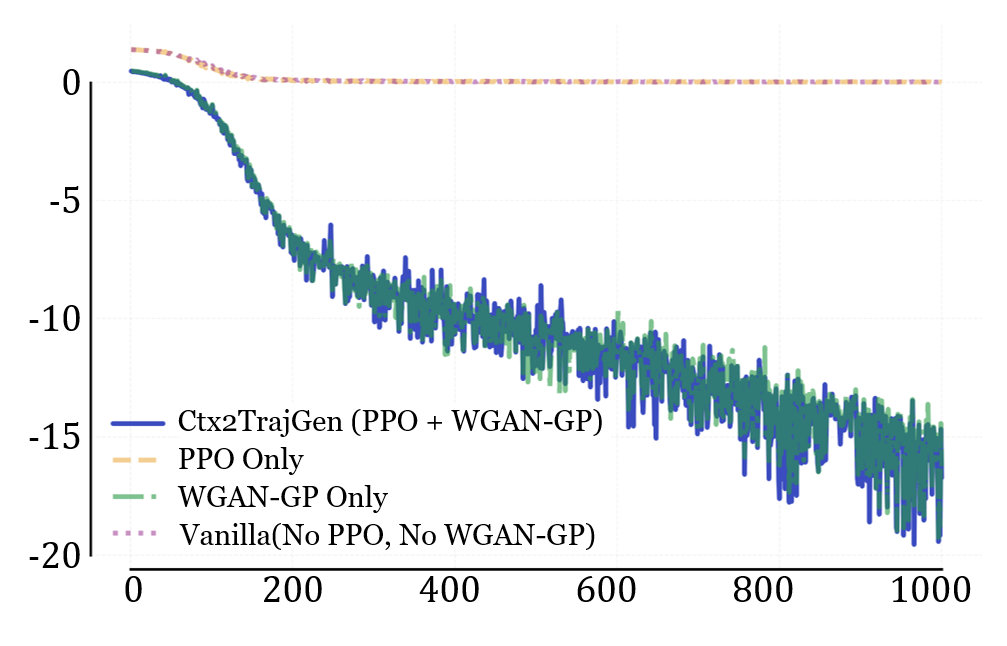}
        \vspace{-0.7cm}
        \caption{Discriminator Loss}\label{fig:disc_loss}
    \end{subfigure}
    
    \begin{subfigure}[b]{0.9\columnwidth}
        \centering
        \includegraphics[width=\linewidth]{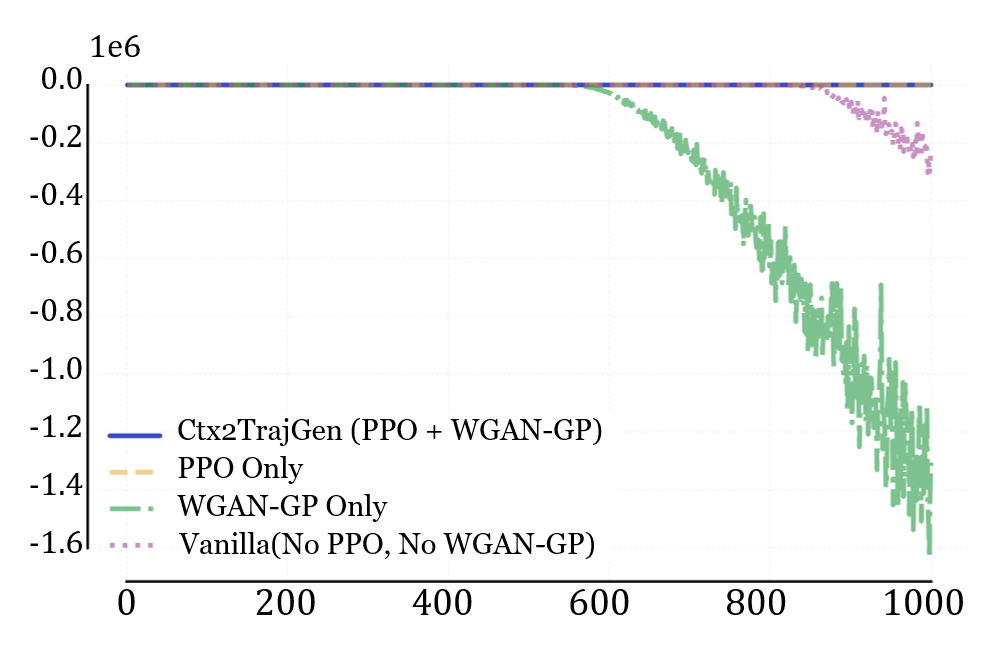}
        \vspace{-0.7cm}
        \caption{Policy Loss}\label{fig:policy_loss}
    \end{subfigure}

    \begin{subfigure}[b]{0.9\columnwidth}
        \centering
        \includegraphics[width=\linewidth]{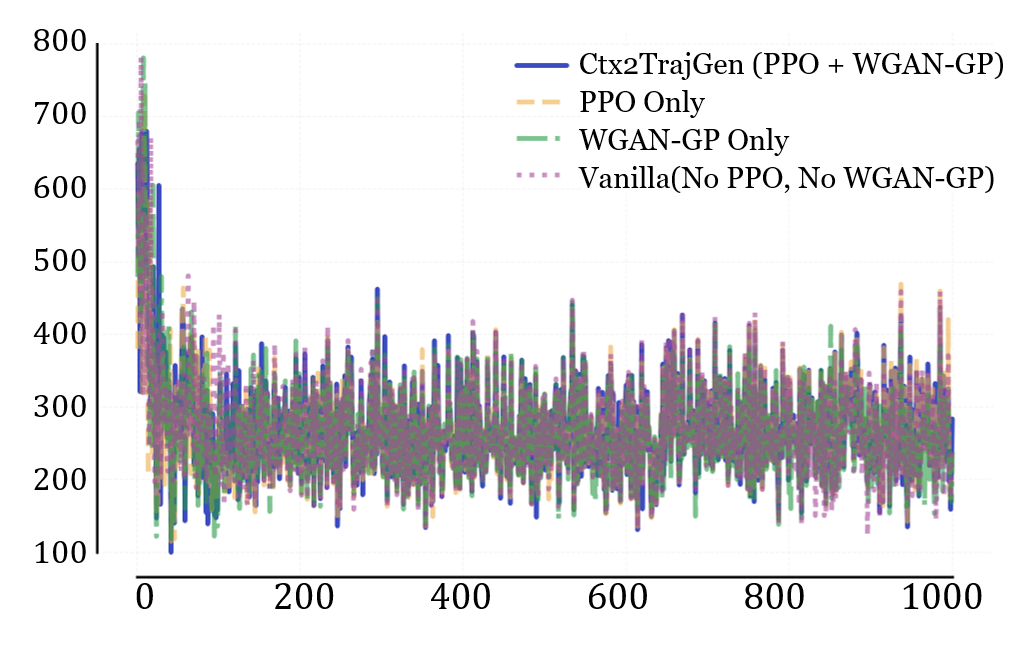}
        \vspace{-0.7cm}
        \caption{Value Loss}\label{fig:value_loss}
    \end{subfigure}

    \caption{Convergence curves of ablation studies evaluating PPO and WGAN-GP stability.}
    \label{fig:fig-ablation-loss}

    \vspace{-5pt}
\end{figure}

Figure~\ref{fig:fig-ablation-loss} provides an overview of learning curve trends across ablation studies of the Ctx2TrajGen framework, assessing the combined PPO + WGAN-GP configuration, PPO-only, WGAN-GP-only, and no PPO and no WGAN-GP configuration to address stability challenges in adversarial imitation learning. In Figure~\ref{fig:disc_loss}, WGAN-GP-inclusive models show a rapid initial decline, followed by a continuous drop toward negative values. This suggests effective learning, as lower values indicate enhanced distribution discrimination. In contrast, WGAN-GP-absent models stay near zero with minimal variation and unstable patterns, highlighting WGAN-GP’s critical role. In Figure~\ref{fig:policy_loss}, PPO-enabled models maintain stability near zero, signifying successful policy optimization, whereas PPO-excluded models exhibit late-stage fluctuations, emphasizing PPO’s necessity. In Figure~\ref{fig:value_loss}, all configurations show a gradual decline with oscillations, with the combined approach offering smoother convergence, suggesting Value Network independence enhanced by integrated optimization. These trends confirm that PPO and WGAN-GP synergistically improve stability and trajectory fidelity, mitigating mode collapse.

\begin{table}[!t]
  \centering
  \renewcommand{\arraystretch}{1.1}
  \fontsize{9pt}{11pt}\selectfont
  \setlength{\tabcolsep}{3.5pt}
  \begin{tabular}{@{} c c | c c c c c @{}}
    \toprule
    \textbf{PPO} & \textbf{WGAN-GP} & \textbf{MMD ↓} & \textbf{WD ↓} & \textbf{KL ↓} & \textbf{JS ↓} & \textbf{Time (s)}\\
    \midrule
    \(\times\)   & \(\times\)   & 0.0130 & 0.4003 & 1.5178 & 0.2820 & 30.5\\
    \checkmark & \(\times\)   & 0.0125 & 0.2977 & 1.4675 & 0.2976 & 31.2\\
    \(\times\)   & \checkmark & 0.0133 & 0.3142 & 2.7538 & 0.3132 & 30.8\\
    \checkmark & \checkmark & \textbf{0.0021} & \textbf{0.2781} & \textbf{1.2543} & \textbf{0.2746} & \textbf{29.8}\\
    \bottomrule
  \end{tabular}
  \caption{Ablation study of Ctx2TrajGen across diverse PPO and WGAN-GP configurations. The check mark (\checkmark) indicates that the corresponding component is included, while the cross (\(\times\)) denotes its exclusion.}
  \label{tab:tab-ablation-study}
  \vspace{-8pt}
\end{table}

The ablation study presented in Table~\ref{tab:tab-ablation-study} quantifies the impact of PPO and WGAN-GP on the Ctx2TrajGen framework’s performance, evaluated through metrics MMD, WD, KL, JS, and computation time. The combined PPO + WGAN-GP configuration achieves the lowest values across MMD, WD, KL, and JS, indicating superior trajectory distribution fidelity and reduced divergence from expert data. This improvement arises from WGAN-GP’s enhanced distribution discrimination, which stabilizes training by minimizing Wasserstein distance, and PPO’s robust policy optimization, which prevents late-stage fluctuations and ensures efficient updates. Individually, WGAN-GP reduces WD and KL, while PPO lowers MMD and JS, but their synergy in the combined setup yields the best overall alignment. Additionally, the combined configuration records the shortest computation time, likely due to stabilized learning reducing iterative adjustments, contrasting with higher times for unstable configurations. These results reinforce that PPO and WGAN-GP collaboratively enhance performance and efficiency, validating the framework’s optimized design.

\subsection{Trajectory-Level Validation}

\begin{figure}[!t]
    \centering
    \includegraphics[width=1.0\linewidth]{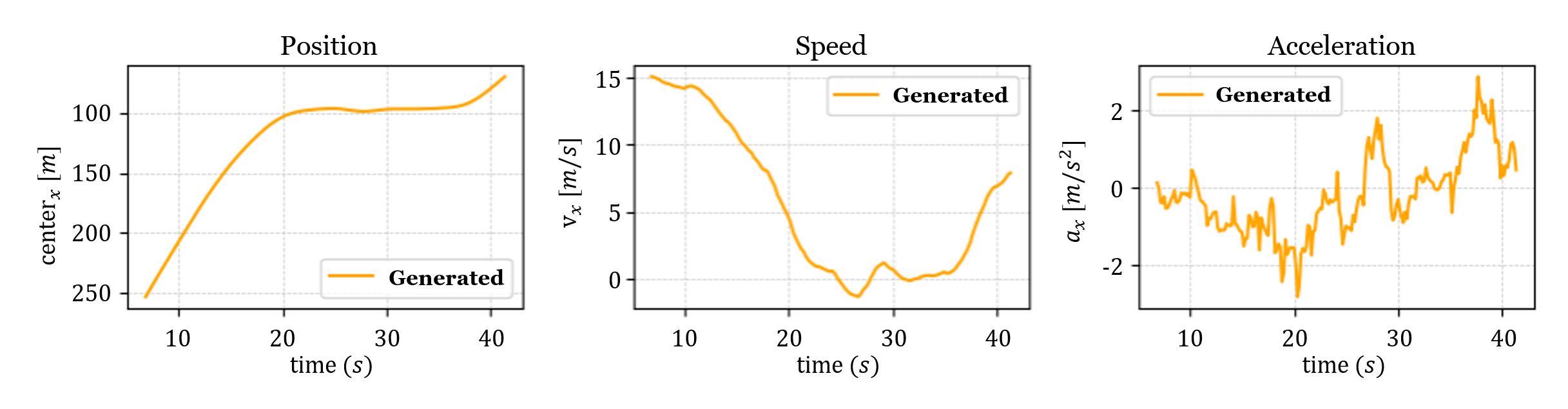}    
    \caption{Temporal profiles of position, speed, and acceleration of an example generated trajectory.}
    \label{fig:fig-trajectory-xva}
\end{figure}

Fig.~\ref{fig:fig-trajectory-xva} illustrates a generated trajectory, where the vehicle exhibits deceleration followed by acceleration, reflecting typical stop-and-go behavior in traffic flow. This pattern, a key phenomenon, includes minor speed fluctuations during standstill phases, indicating a deviation from zero speed. This observation suggests a potential area for model refinement to enhance precision in static conditions.

\begin{figure}[!t]
    \centering
    \begin{subfigure}[b]{1.0\columnwidth}
        \centering
        \includegraphics[width=\linewidth]{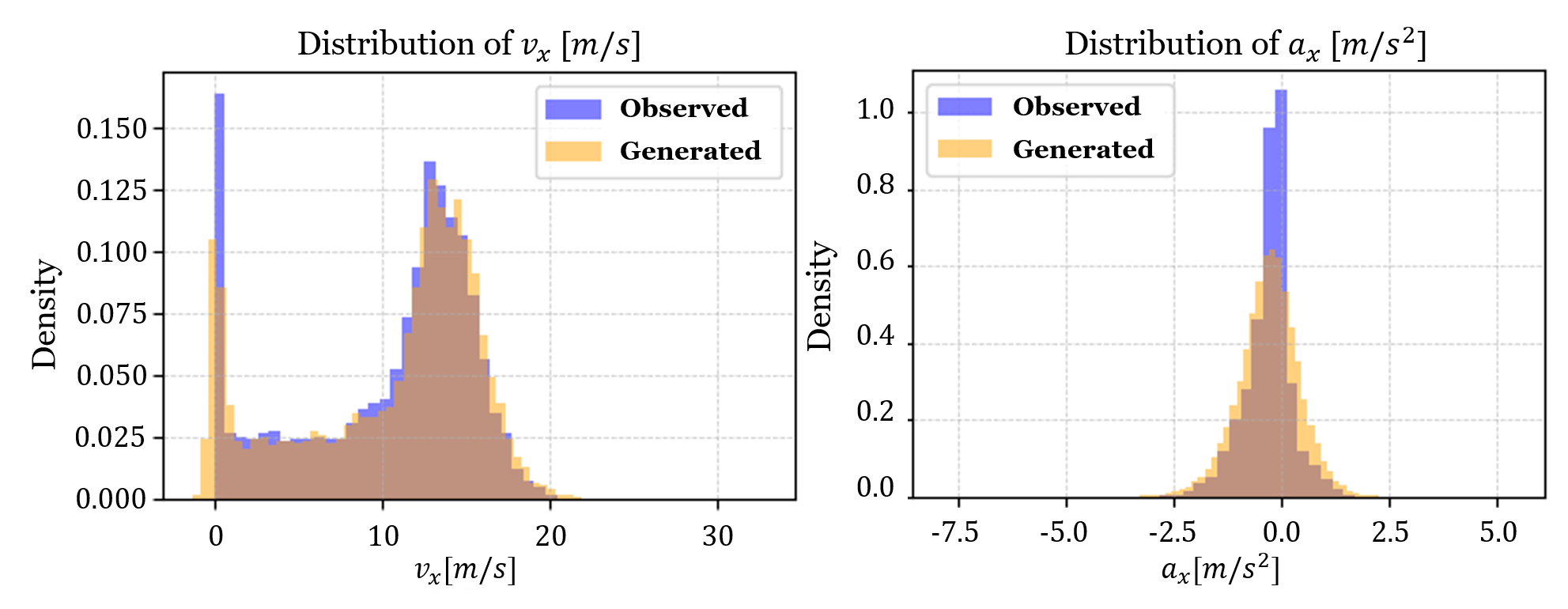}
        \caption{Distribution of speed and acceleration comparing observed and generated data}\label{fig:traj_distr}
    \end{subfigure}

    \vspace{0.1cm}

    \begin{subfigure}[b]{1.0\columnwidth}
        \centering
        \includegraphics[width=\linewidth]{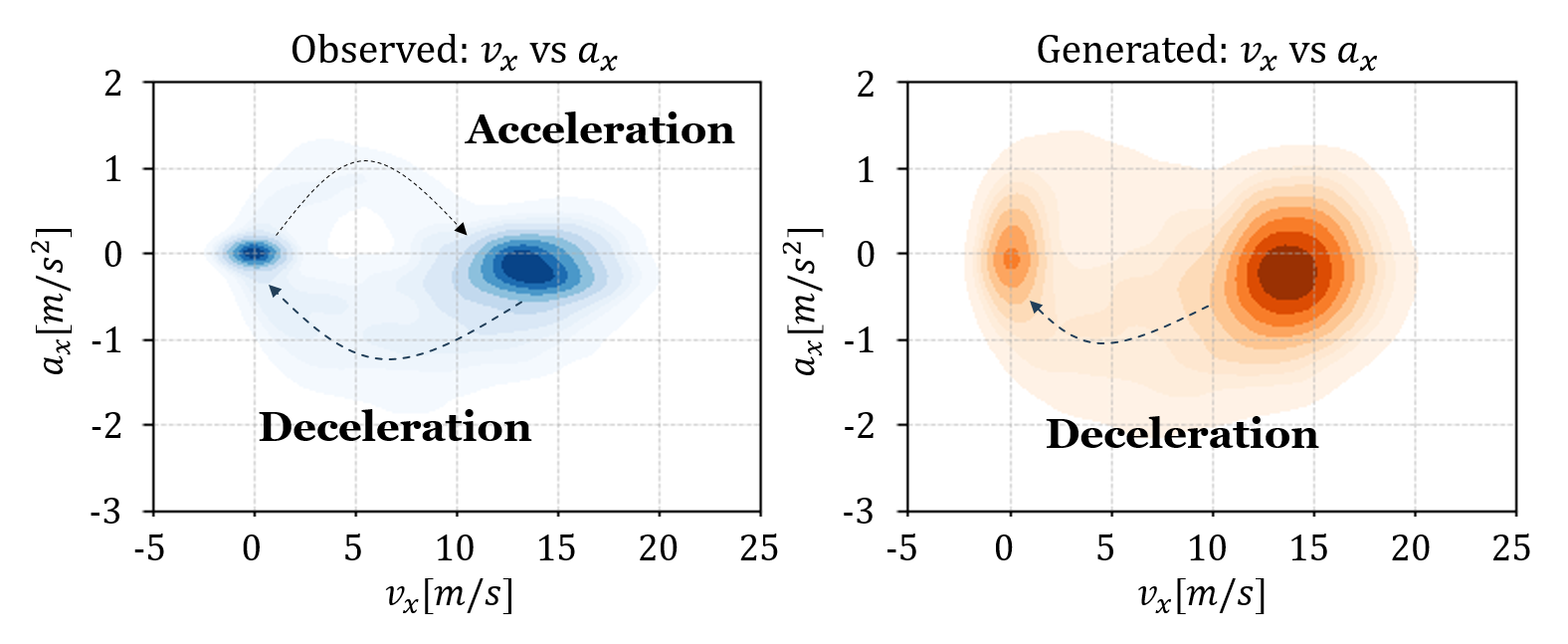}
        \caption{KDE plots comparing speed and acceleration for observed and generated data}\label{fig:traj_scatter}
    \end{subfigure}

    \caption{Comparative analysis of trajectory-level similarity between observed and generated data.}
    \label{fig:fig-trajectory-level-similarity-validation}
\end{figure}

Figure~\ref{fig:traj_distr} presents the distributions of speed and acceleration for both observed and generated data. The generated data exhibits a distribution broadly comparable to the observed data. However, the presence of negative speed values is noted in the generated data, and while the observed data demonstrates a pronounced concentration of acceleration values around zero, the generated data reveals a somewhat wider spread, indicating a tendency of the model to produce subtle variations during near-stop conditions. Figure~\ref{fig:traj_scatter} displays kernel density estimation (KDE) plots comparing speed and acceleration for the observed and generated datasets. The observed data distinctly delineates acceleration and deceleration patterns, which the generated data replicates in its deceleration trend, though acceleration from standstill appears less prominently represented, potentially due to limited exposure to such scenarios in the training data.


\section{Conclusion}This study introduces the Ctx2TrajGen framework, a GAIL-based model that generates microscopic trajectories by imitating expert driving patterns derived from the real-world DRIFT dataset, integrating PPO for stable policy optimization and WGAN-GP for enhanced discriminator learning. This approach effectively leverages dynamic inputs from surrounding vehicles and lane curvature data, enabling context-sensitive and interaction-aware trajectory generation across diverse traffic scenarios. Experimental results highlight Ctx2TrajGen’s superior performance over baseline methods across MMD, KL, and JS metrics, with significant improvements in distribution fidelity and reduced mode collapse risks. Unlike single-metric-focused approaches, it achieves a balanced synthesis of trajectory diversity and consistency, supporting realistic generation within complex road structures and dynamic environments. The framework establishes a robust foundation for autonomous driving simulation and traffic analysis, with its extensible architecture facilitating future research into adaptive driver behaviors and long-term sequence modeling.



\end{document}